\documentclass[11pt]{article}
\usepackage[a4paper, margin=1in]{geometry}
\usepackage[usenames,dvipsnames,svgnames,table]{xcolor} 

\usepackage{algorithm}
\usepackage{algorithmic}
\floatname{algorithm}{Algorithm}

\usepackage{longtable}
\usepackage{hhline}
\usepackage{float}

\usepackage{enumitem} 
\usepackage{rotfloat} 
\usepackage{graphicx}
\usepackage{amssymb, amsmath, amsthm}
\usepackage{tikz}
\usetikzlibrary{3d, calc}
\usepackage{verbatim}
\usepackage{natbib}
\usepackage{hyperref}
\usepackage{authblk}
\usepackage{multirow}
\usepackage{subcaption} 
\usepackage{array}
\newcolumntype{L}{>{\centering\arraybackslash}m{0.1\linewidth}}

\begin{document}
	
\title{Constant Metric Scaling in Riemannian Computation}
\author[1]{Kisung You}
\affil[1]{Department of Mathematics, Baruch College}
\date{}

\maketitle
\begin{abstract}
Constant rescaling of a Riemannian metric appears in many computational settings,
often through a global scale parameter that is introduced either explicitly or
implicitly. Although this operation is elementary, its consequences are not
always made clear in practice and may be confused with changes in curvature,
manifold structure, or coordinate representation. In this note, we provide a
short, self-contained account of constant metric scaling on arbitrary Riemannian
manifolds. We distinguish between quantities that change under such a scaling,
including norms, distances, volume elements, and gradient magnitudes, and
geometric objects that remain invariant, such as the Levi--Civita connection,
geodesics, exponential and logarithmic maps, and parallel transport. We also
discuss implications for Riemannian optimization, where constant metric scaling
can be interpreted as a global rescaling of step sizes rather than a
modification of the underlying geometry. The goal of this note is purely
expository and is intended to clarify how a global metric scale parameter can be
introduced in Riemannian computation without altering the geometric structures
on which these methods rely.
\end{abstract}

\section{Introduction}

Riemannian geometry has become an increasingly common tool in modern computational settings, including optimization on manifolds \citep{boumal_2023_IntroductionOptimizationSmooth} and geometric data analysis \citep{bhattacharya_2012_NonparametricInferenceManifolds, patrangenaru_2016_NonparametricStatisticsManifolds}. In many of these applications, one can encounter  a global parameter that controls the overall scale of the Riemannian metric. Such a parameter is often introduced implicitly through a temperature, normalization constant, or so-called curvature parameter, and is sometimes interpreted as modifying the geometric structure of the underlying space. While this intuition is natural, it can obscure a simple but important distinction: when a Riemannian metric is rescaled by a constant factor, the resulting transformation alters measurements but leaves the underlying geometry unchanged.

The purpose of this note is to provide a self-contained account of constant metric scaling on Riemannian manifolds and to spell out its consequences for Riemannian computation. The geometric facts themselves are classical and well known in differential geometry \citep{docarmo_1992_RiemannianGeometry, lee_1997_RiemannianManifoldsIntroduction}. However, their implications are not always made explicit in computational or modeling contexts, where metric scaling is frequently conflated with changes in curvature, manifold constraints, or coordinate representations. In particular, it is not uncommon for constant metric rescaling to be treated as if it modifies geodesics, exponential maps, or even the manifold itself. One motivation for this paper is to disentangle these ideas and to record what does and does not change under constant metric scaling in one place.

We focus on a simple but practically relevant setting. Given a Riemannian manifold $(M,g)$, we consider a rescaled metric $\tilde g = \lambda g$ with $\lambda>0$ constant. Under this transformation, numerical quantities such as norms, distances, volumes, and gradient magnitudes are rescaled in a predictable way. At the same time, the Levi--Civita connection, geodesics, exponential and logarithmic maps, and parallel transport remain exactly the same. As a consequence, many computational tools used in practice, particularly those based on geodesic constructions, are invariant under constant metric scaling and require no modification. From a practical view, the effect of metric scaling can often be interpreted as a global rescaling of step sizes rather than a change in geometric structure.

This paper is intentionally expository. We do not aim to introduce new theory or algorithms, but rather to give explicit derivations and a systematic classification of invariant and variant quantities under constant metric scaling. The presentation is kept general, applying to arbitrary Riemannian manifolds, and is meant to serve as a compact reference for readers who wish to incorporate a flexible metric scale parameter into Riemannian models or algorithms without inadvertently altering the geometry they rely on. Throughout, we adopt a deliberately modest perspective and the results collected here may be viewed as a small exercise in clarifying familiar geometric facts, motivated by their recurring appearance in contemporary computational practice.

\section{Setup and Notation}

Let $M$ be a smooth manifold of dimension $n$, equipped with a Riemannian metric $g$. For each point $p \in M$, the metric $g_p$ induces an inner product on the tangent space $T_p M$, and we write $\|\cdot\|_g$ for the associated norm. The Riemannian distance induced by $g$ is denoted by $d_g$, and the corresponding Riemannian volume form by $d\mathrm{vol}_g$.

Throughout the paper, we consider a \textit{constant rescaling} of the metric $g$. That is, for a fixed constant $\lambda > 0$, we define a new Riemannian metric
\begin{equation*}
    \tilde g := \lambda g.
\end{equation*}
All geometric objects associated with the scaled metric will be denoted with a tilde when needed. Our interest is in understanding how basic geometric and computational quantities change or remain unchanged under this transformation. We emphasize that the underlying smooth manifold $M$ is unchanged, and that the scaling factor $\lambda$ is constant over the manifold. No assumptions are made on curvature, topology, or global structure, beyond standard smoothness conditions ensuring that the usual constructions of Riemannian geometry are well defined.

\subsection{Tangent spaces and basic constructions}

For each $p \in M$, the tangent space $T_p M$ is the same whether it is viewed as part of $(M,g)$ or $(M,\tilde g)$. The difference lies only in how vectors in $T_p M$ are measured. In particular,
\begin{equation*}
\|v\|_{\tilde g}^2 = \tilde g_p(v,v) = \lambda g_p(v,v)
\quad \text{for all } v \in T_p M.
\end{equation*}
We will make repeated use of this elementary relation in what follows.

The Levi--Civita connection associated with $g$ is denoted by $\nabla$, and the one associated with $\tilde g$ by $\tilde\nabla$. As is standard, geodesics are defined as curves $\gamma : I \to M$ satisfying the geodesic equation
\begin{equation*}
\nabla_{\dot\gamma} \dot\gamma = 0,
\end{equation*}
and similarly for $\tilde\nabla$. The relationship between these two connections under constant metric scaling will play a central role in the subsequent sections.

\subsection{Geodesic-based maps}

Given a point $p \in M$, the exponential map associated with $g$ is denoted by
\begin{equation*}
    \exp_p^g : T_p M \to M,
\end{equation*}
defined by $\exp_p^g(v) = \gamma_v(1)$, where $\gamma_v$ is the unique geodesic satisfying $\gamma_v(0) = p$ and $\dot\gamma_v(0) = v$. When the metric is clear from context, we will simply write $\exp_p$.

Whenever the exponential map is locally invertible, we denote its local inverse by the logarithm map
\begin{equation*}
    \log_p^g : M \supset U \to T_p M,
\end{equation*}
again omitting the superscript when no confusion arises. Parallel transport along a smooth curve $\gamma$ with respect to the metric $g$ is denoted by $P_\gamma^g$, and is defined in the usual way via covariant differentiation along $\gamma$. All of these constructions have direct analogues for the scaled metric $\tilde g$. One of the main points of this paper is to clarify which of these objects depend on the choice of metric scale and which do not.

\subsection{Gradients and optimization}

For a smooth function $f : M \to \mathbb{R}$, the Riemannian gradient with respect to $g$, denoted $\nabla_g f$, is defined by the identity
\begin{equation*}
g(\nabla_g f, X) = df(X)
\quad \text{for all vector fields } X.
\end{equation*}
The gradient with respect to the scaled metric $\tilde g$ is defined analogously and denoted by $\nabla_{\tilde g} f$. Since gradient-based methods are central to many computational applications, understanding how $\nabla_g f$ and $\nabla_{\tilde g} f$ are related under constant metric scaling will be one of our main concerns in later sections.

\section{Quantities That Change Under Constant Metric Scaling}

We begin by describing those quantities that are directly affected by constant rescaling of the Riemannian metric. Throughout this section, $(M,g)$ denotes a Riemannian manifold, and $\tilde g = \lambda g$ with $\lambda > 0$ denotes the scaled metric introduced in Section~2. Our goal is not to present new results, but to record explicitly how basic measurements transform under this scaling.

\subsection{Norms of tangent vectors}

Let $p \in M$ and let $v \in T_p M$. By definition, the norm of $v$ with respect to the metric $g$ is given by
\begin{equation*}
\|v\|_g^2 := g_p(v,v),    
\end{equation*}
and similarly, the norm with respect to the scaled metric $\tilde g$ is
\begin{equation*}
    \|v\|_{\tilde g}^2 := \tilde g_p(v,v).
\end{equation*}
Since $\tilde g = \lambda g$, we have
\begin{equation*}
\|v\|_{\tilde g}^2
= \lambda g_p(v,v)
= \lambda \|v\|_g^2.    
\end{equation*}
Taking square roots yields
\begin{equation*}
    \|v\|_{\tilde g} = \sqrt{\lambda}\,\|v\|_g.
\end{equation*}
Thus, constant metric scaling rescales the norm of every tangent vector by the same factor $\sqrt{\lambda}$. No directional information is altered; only the overall magnitude of vectors is affected.

\subsection{Lengths of curves and Riemannian distance}

The effect of metric scaling on curve lengths follows directly from the scaling
of tangent vector norms. Let $\gamma : [0,1] \to M$ be a smooth curve. The length
of $\gamma$ with respect to the metric $g$ is defined by
\[
L_g(\gamma)
= \int_0^1 \|\dot\gamma(t)\|_g \, dt.
\]
Under the scaled metric $\tilde g$, the length becomes
\[
L_{\tilde g}(\gamma)
= \int_0^1 \|\dot\gamma(t)\|_{\tilde g} \, dt
= \int_0^1 \sqrt{\lambda}\,\|\dot\gamma(t)\|_g \, dt
= \sqrt{\lambda}\, L_g(\gamma).
\]
In particular, the length of every curve is multiplied by the same factor
$\sqrt{\lambda}$. As a consequence, the Riemannian distance induced by $\tilde g$
satisfies
\[
d_{\tilde g}(p,q)
= \inf_{\gamma \in \gamma_{\overline{pq}}} L_{\tilde g}(\gamma)
= \sqrt{\lambda}\,
\inf_{\gamma \in \gamma_{\overline{pq}}} L_g(\gamma)
= \sqrt{\lambda}\, d_g(p,q),
\]
for $\gamma_{\overline{pq}} = \lbrace\gamma~\vert~\gamma(0)=p, \; \gamma(1)=q\rbrace$.

Importantly, while the numerical value of the distance changes, the curves that
minimize length do not. Any minimizing geodesic for $d_g$ is also a minimizing
geodesic for $d_{\tilde g}$, and vice versa. In this sense, constant metric
scaling rescales distances without altering the underlying geodesic paths.

\subsection{Riemannian volume}

We next consider how the Riemannian volume form changes under constant metric
scaling. Let $\dim M = n$, and let $(x^1,\dots,x^n)$ be a local coordinate chart
on $M$. In these coordinates, the metric $g$ is represented by a symmetric
positive-definite matrix $(g_{ij})$, and the associated Riemannian volume form is
given by
\[
d\mathrm{vol}_g
= \sqrt{\det(g_{ij})}\, dx^1 \cdots dx^n.
\]
Under the scaled metric $\tilde g = \lambda g$, the coordinate representation
becomes $\tilde g_{ij} = \lambda g_{ij}$. Since multiplying a matrix by
$\lambda$ scales its determinant by $\lambda^n$, we obtain
\[
\det(\tilde g_{ij})
= \lambda^n \det(g_{ij}).
\]
Taking square roots yields
\[
\sqrt{\det(\tilde g_{ij})}
= \lambda^{n/2} \sqrt{\det(g_{ij})}.
\]
Consequently, the Riemannian volume form satisfies
\[
d\mathrm{vol}_{\tilde g}
= \lambda^{n/2}\, d\mathrm{vol}_g.
\]

Thus, constant metric scaling rescales volume by a factor depending only on the
dimension of the manifold. While this observation is elementary, it is relevant
in applications involving integration or probability densities on manifolds, as
it makes explicit how normalization constants depend on the metric scale.

\subsection{Gradient of a function}

Finally, we examine the effect of constant metric scaling on the Riemannian
gradient of a function. Let $f : M \to \mathbb{R}$ be a smooth function. The
Riemannian gradient of $f$ with respect to the metric $g$, denoted $\nabla_g f$,
is defined by the identity
\[
g(\nabla_g f, X) = df(X)
\quad \text{for all vector fields } X.
\]
Similarly, the gradient with respect to the scaled metric $\tilde g$ is defined
by
\[
\tilde g(\nabla_{\tilde g} f, X) = df(X)
\quad \text{for all vector fields } X.
\]
Using the relation $\tilde g = \lambda g$, we may rewrite the second identity as
\[
\lambda g(\nabla_{\tilde g} f, X) = df(X).
\]
Comparing this with the defining equation for $\nabla_g f$, we see that
\[
g(\nabla_g f, X) = \lambda g(\nabla_{\tilde g} f, X)
\quad \text{for all } X,
\]
which implies
\[
\nabla_{\tilde g} f = \lambda^{-1} \nabla_g f.
\]

Therefore, constant metric scaling rescales the Riemannian gradient by the reciprocal
factor $1/\lambda$. From a computational perspective, this means that gradient
directions are unchanged, while their magnitudes are uniformly rescaled. In many
optimization algorithms on manifolds, this effect can be interpreted as a global
adjustment of step sizes rather than a modification of the underlying geometry.

\section{Quantities That Do Not Change}

We now turn to those geometric objects that remain invariant under constant
metric scaling. While the numerical measurements discussed in Section~3 are
rescaled, the underlying geometric structures governing geodesic-based
computation are unaffected. This invariance is central to the practical
consequences of constant metric scaling.

\subsection{Levi--Civita connection}

Let $\nabla$ denote the Levi--Civita connection associated with the metric $g$,
and let $\tilde\nabla$ denote the Levi--Civita connection associated with the
scaled metric $\tilde g = \lambda g$. Recall that the Levi--Civita connection is
uniquely characterized by two properties: it is torsion-free and compatible with
the metric.

By construction, $\nabla$ is torsion-free. So it remains torsion-free
independently of the choice of metric. To verify metric compatibility with
respect to $\tilde g$, let $X,Y,Z$ be smooth vector fields on $M$. Using the
relation $\tilde g = \lambda g$ and the fact that $\lambda$ is constant, we
compute
\[
X\bigl(\tilde g(Y,Z)\bigr)
= X\bigl(\lambda g(Y,Z)\bigr)
= \lambda X\bigl(g(Y,Z)\bigr).
\]
Because $\nabla$ is compatible with $g$, we have
\[
X\bigl(g(Y,Z)\bigr)
= g(\nabla_X Y, Z) + g(Y, \nabla_X Z).
\]
Combining these expressions yields
\[
X\bigl(\tilde g(Y,Z)\bigr)
= \tilde g(\nabla_X Y, Z) + \tilde g(Y, \nabla_X Z).
\]
Thus, $\nabla$ is also compatible with the metric $\tilde g$. By uniqueness of
the Levi--Civita connection, it follows that
\[
\tilde\nabla = \nabla.
\]

\subsection{Geodesics}

Geodesics are defined as curves $\gamma : I \to M$ satisfying the geodesic
equation
\[
\nabla_{\dot\gamma} \dot\gamma = 0.
\]
When the metric is changed from $g$ to $\tilde g$, the geodesic equation is
defined using the corresponding Levi--Civita connection $\tilde\nabla$. Since
$\tilde\nabla = \nabla$, the geodesic equations for $g$ and $\tilde g$ coincide.

As a result, a curve is a geodesic with respect to $g$ if and only if it is a
geodesic with respect to $\tilde g$. In particular, given identical initial
conditions $\gamma(0) = p$ and $\dot\gamma(0) = v$, the resulting geodesic curve
is the same for both metrics. Constant metric scaling therefore does not alter
the set of geodesic curves or their parametrization by initial velocity.

\subsection{Exponential and logarithm maps}

Fix a point $p \in M$. The exponential map associated with the metric $g$ is
defined by
\[
\exp_p^g(v) = \gamma_v(1),
\]
where $\gamma_v$ is the unique geodesic satisfying $\gamma_v(0) = p$ and
$\dot\gamma_v(0) = v$. The exponential map for the scaled metric $\tilde g$ is
defined analogously using $\tilde\nabla$. Since the geodesics determined by $g$ and $\tilde g$ coincide for identical
initial conditions, it follows immediately that
\[
\exp_p^{\tilde g}(v) = \exp_p^g(v)
\quad \text{for all } v \in T_p M.
\]
Thus, constant metric scaling does not affect the exponential map as a mapping
from the tangent space to the manifold. The only difference lies in the
interpretation of the norm of the tangent vector $v$, which is rescaled as
discussed in Section~3.

Whenever the exponential map $\exp_p^g$ is locally invertible, its local inverse
is given by the logarithm map $\log_p^g$. Since the exponential maps associated
with $g$ and $\tilde g$ coincide, their local inverses also coincide on common
domains of definition. In particular,
\[
\log_p^{\tilde g}(q) = \log_p^g(q)
\]
whenever both sides are well defined. Although the tangent vector returned by the logarithm map is the same for both
metrics, its norm depends on the metric used for measurement. Specifically,
\[
\|\log_p^{\tilde g}(q)\|_{\tilde g}
= \sqrt{\lambda}\,\|\log_p^g(q)\|_g,
\]
which is consistent with the scaling of Riemannian distance established before.

\subsection{Parallel transport}

Let $\gamma : [0,1] \to M$ be a smooth curve. Parallel transport of a tangent
vector along $\gamma$ with respect to the metric $g$ is defined by solving the
ordinary differential equation
\[
\nabla_{\dot\gamma(t)} V(t) = 0,
\qquad V(0) = v_0,
\]
where $v_0 \in T_{\gamma(0)} M$. The parallel transport map is then given by
$P_\gamma^g(v_0) = V(1)$.

Since the Levi--Civita connections associated with $g$ and $\tilde g$ coincide,
the defining differential equation for parallel transport is identical under
both metrics. Consequently, the parallel transport maps agree:
\[
P_\gamma^{\tilde g} = P_\gamma^g.
\]
As in the previous cases, constant metric scaling affects only the metric used to
measure transported vectors, not the transport operation itself.

\section{Implications for Riemannian Computation}

The aforementioned results have several direct implications for
computational procedures on Riemannian manifolds. In this section, we briefly
discuss how constant metric scaling interacts with common algorithmic
constructions, with an emphasis on gradient-based methods and modeling
considerations. The discussion is intended to be conceptual rather than
algorithm-specific.

\subsection{Riemannian optimization algorithms}

Many optimization algorithms on manifolds are based on repeated evaluations of
Riemannian gradients followed by updates defined through the exponential map or
a suitable retraction. A prototypical update of Riemannian gradient descent
takes the form
\[
x_{k+1} = \exp_{x_k}\bigl(-\eta \, \nabla_g f(x_k)\bigr),
\]
where $\eta > 0$ is a step size and $\nabla_g f$ denotes the Riemannian gradient
with respect to the metric $g$.

When the metric is replaced by the scaled metric $\tilde g = \lambda g$, the
corresponding gradient satisfies $\nabla_{\tilde g} f = \lambda^{-1}
\nabla_g f$, while the exponential map remains unchanged. As a result, the
update becomes
\[
x_{k+1}
= \exp_{x_k}\bigl(-\eta \, \nabla_{\tilde g} f(x_k)\bigr)
= \exp_{x_k}\bigl(-\eta \lambda^{-1} \nabla_g f(x_k)\bigr).
\]
Thus, from the perspective of the update rule, constant metric scaling is
equivalent to a rescaling of the step size by a factor $\lambda^{-1}$. The
direction of the update is unaffected, and no modification of the exponential
map is required.

This observation extends immediately to other first-order methods, including
stochastic gradient descent and related variants. In such settings, constant
metric scaling influences the magnitude of updates but does not alter their
geometric interpretation.

\subsection{Joint optimization over points and metric scale}

In some modeling or algorithmic settings, the metric scale parameter $\lambda$
may itself be treated as a tunable or learnable quantity. The results above
suggest that, when $\lambda$ enters only through a constant rescaling of the
metric, its effect is largely decoupled from the geometric aspects of the
optimization.

Specifically, updates of the form
\[
x_{k+1} = \exp_{x_k}\bigl(-\eta \, \nabla_{\tilde g} f(x_k)\bigr)
\]
depend on $\lambda$ only through the overall magnitude of the gradient. This
decoupling allows one to interpret $\lambda$ as a global measurement scale,
whose adjustment does not require changes to the manifold-valued variables or
to the geodesic-based computational machinery.

From this viewpoint, optimizing over $\lambda$ can often be understood as
controlling the effective step size or sensitivity of the algorithm, rather
than modifying the underlying geometry on which the computation is performed.

\subsection{Manifold constraints and coordinate representations}

A further consequence of constant metric scaling is that manifold-defining
constraints and coordinate representations are unaffected. Since the smooth
manifold $M$ and its tangent spaces are unchanged, any constraints used to
define $M$ such as normalization constraints in embedded manifolds remain valid
independently of the choice of metric scale.

Particularly, constant metric scaling does not require re-normalization of
coordinates, changes in ambient embeddings, or modifications of projection or
retraction operators. All such constructions depend on the manifold structure
and connection, which are invariant under constant scaling. The metric scale
enters only through the measurement of lengths, distances, and gradients. This separation between geometric structure and metric scale is one of the main
practical advantages of constant metric scaling in computational applications.

\section{Discussion and Conclusion}

This note has examined the effects of constant metric scaling on Riemannian
manifolds, with the goal of clarifying what aspects of geometry and computation
are affected by such a transformation and which are not. Although the underlying
geometric facts are elementary, their implications are not always made explicit
in computational settings, where metric scaling is sometimes conflated with
changes in curvature, manifold structure, or coordinate representation.

Our main observation is that constant rescaling of a Riemannian metric introduces
a global measurement scale without modifying the geometric structures that govern
geodesic-based computation. Quantities such as norms, distances, volumes, and
gradient magnitudes are rescaled in a predictable manner, while the Levi--Civita
connection, geodesics, exponential and logarithmic maps, and parallel transport
remain unchanged. As a result, the core computational tools used in Riemannian
optimization and related algorithms are invariant under constant metric scaling.

From a practical perspective, this invariance implies that existing
implementations of geometric operations need not be altered when a global metric
scale parameter is introduced. In optimization algorithms, the effect of metric
scaling can often be interpreted as a uniform rescaling of step sizes rather than
a modification of the geometry itself. Similarly, manifold-defining constraints
and coordinate representations are independent of the metric scale and therefore
remain valid without adjustment.

It is worth emphasizing that the scope of this paper is deliberately limited to
constant metric scaling. When the scaling factor varies across the manifold, the
situation changes qualitatively: the Levi--Civita connection is no longer
invariant, and geodesic-based constructions are affected in more substantial
ways. Such non-constant conformal transformations raise additional analytical and
computational issues that lie beyond the intent of this note.

The perspective adopted here is intentionally modest. Rather than proposing new
models or algorithms, we have aimed to collect and organize familiar results in a
form that is directly relevant to computation. By making explicit the separation
between geometric structure and metric scale, we hope this note serves as a
useful reference for readers who wish to incorporate a flexible metric scale
parameter into Riemannian models and algorithms while preserving the geometry on
which their computations rely.

\bibliographystyle{dcu}
\bibliography{referencesZotero}

@book{boumal_2023_IntroductionOptimizationSmooth,
	address = {Cambridge New York, NY},
	title = {An {Introduction} to {Optimization} on {Smooth} {Manifolds}},
	isbn = {978-1-009-16616-4},
	doi = {10.1017/9781009166164},
	language = {eng},
	publisher = {Cambridge University Press},
	author = {Boumal, Nicolas},
	year = {2023},
}

@book{patrangenaru_2016_NonparametricStatisticsManifolds,
	address = {Boca Raton},
	title = {Nonparametric {Statistics} on {Manifolds} and {Their} {Applications} to {Object} {Data} {Analysis}},
	isbn = {978-1-4398-2050-6},
	publisher = {CRC Press, Taylor \& Francis Group},
	author = {Patrangenaru, Victor and Ellingson, Leif},
	year = {2016},
}

@book{lee_1997_RiemannianManifoldsIntroduction,
	address = {New York},
	series = {Graduate {Texts} in {Mathematics}},
	title = {Riemannian {Manifolds}: {An} {Introduction} to {Curvature}},
	volume = {176},
	isbn = {978-0-387-98271-7 978-0-387-98322-6},
	shorttitle = {Riemannian manifolds},
	publisher = {Springer},
	author = {Lee, John M.},
	year = {1997},
}

@book{docarmo_1992_RiemannianGeometry,
	address = {Boston},
	series = {Mathematics. {Theory} \& {Applications}},
	title = {Riemannian {Geometry}},
	isbn = {978-0-8176-3490-2 978-3-7643-3490-1},
	language = {eng},
	publisher = {Birkhäuser},
	author = {do Carmo, Manfredo Perdigão},
	year = {1992},
}

@book{bhattacharya_2012_NonparametricInferenceManifolds,
	address = {Cambridge},
	title = {Nonparametric {Inference} on {Manifolds}: {With} {Applications} to {Shape} {Spaces}},
	isbn = {978-1-139-09476-4},
	shorttitle = {Nonparametric {Inference} on {Manifolds}},
	url = {http://ebooks.cambridge.org/ref/id/CBO9781139094764},
	doi = {10.1017/CBO9781139094764},
	urldate = {2021-06-11},
	publisher = {Cambridge University Press},
	author = {Bhattacharya, Abhishek and Bhattacharya, Rabi},
	year = {2012},
}


\end{document}